\pdfoutput=1
\documentclass[11pt]{article}
\usepackage{multirow}
\usepackage[table,xcdraw]{xcolor}

\usepackage[]{emnlp2021}

\usepackage{times}
\usepackage{latexsym}
\usepackage{comment}
\usepackage{times}
\usepackage{latexsym}
\usepackage{xspace}

\usepackage[T1]{fontenc}
\usepackage[utf8]{inputenc}

\usepackage{microtype}

\usepackage{amsmath}
\usepackage{amsfonts}
\usepackage{graphicx,subfigure}
\usepackage{ctable}
\usepackage{comment}
\usepackage{multicol}
\usepackage{multirow}
\usepackage{epsfig}
\usepackage{pifont}
\usepackage{dsfont}

\usepackage{amssymb}% http://ctan.org/pkg/amssymb
\usepackage{pifont}% http://ctan.org/pkg/pifont
\newcommand{\cmark}{\ding{51}}%
\usepackage[T1]{fontenc}
\usepackage[utf8]{inputenc}

\usepackage{microtype}

\title{\dataset: Generating Captions for Scientific Figures}

\newcommand{\kenneth}[1]{}
\newcommand{\edward}[1]{}
\newcommand{\lee}[1]{}
\newcommand{\cy}[1]{}

\newcommand{\dataset}{\textsc{SciCap}\xspace}

\author{Ting-Yao (Edward) Hsu,~~C. Lee Giles,~~Ting-Hao `Kenneth' Huang\\
  Pennsylvania State University\\University Park, PA, USA \\
  \texttt{\{txh357,~clg20,~txh710\}@psu.edu} 
  }

\begin{document}
\maketitle

\begin{abstract}

%edited by MB
Researchers use figures to communicate rich, complex information in scientific papers.
The captions of these figures are critical to conveying effective messages. 
However, low-quality figure captions commonly occur in scientific articles and may decrease understanding.
In this paper, we propose an end-to-end neural framework to automatically generate informative, high-quality captions for scientific figures.
%When researchers write papers, a neural-caption model trained on captions can be used to suggest useful captions.
To this end, we introduce \textbf{\dataset},\footnote{\dataset is available at: \url{https://github.com/tingyaohsu/SciCap}} a large-scale figure-caption dataset based on computer science arXiv papers published between 2010 and 2020.
After pre-processing -- including figure-type classification, sub-figure identification, text normalization, and caption text selection -- \dataset contained more than two million figures extracted from over 290,000 papers.
We then established baseline models that caption graph plots, the dominant (19.2\%) figure type.
The experimental results 
%We also established several image-captioning baselines, 
showed both opportunities and steep challenges of generating captions for scientific figures.

\end{abstract}

% Our Google Doc: https://docs.google.com/document/d/1zwVm75HjeVSQJTapYgGR8FIlEVJiDRBKcyPWspLeXxI/edit?usp=sharing

% Rebuttal: https://docs.google.com/document/d/1eyrbN5oGtA-vUNX1evi1PhjADMxcqZFwfDgSzhyCB1g/edit

\section{Introduction}
\label{sec:introduction}
% Kenneth: I did a sentence-by-sentence re-write based on your text. Try to use simple, direct, short sentences in your future writing.

%\lee{Hi Lee, you can use \lee{} to add comments! }

%\kenneth{(1) We will need to pick one or two examples from the data set. I want a few examples of Graph Plot (image) and show their captions (text) that go through Step 1, 2, ..., 6. Showing how we processed/normalized the data.}

\begin{figure}[t]
\centering
\includegraphics[width=\columnwidth]{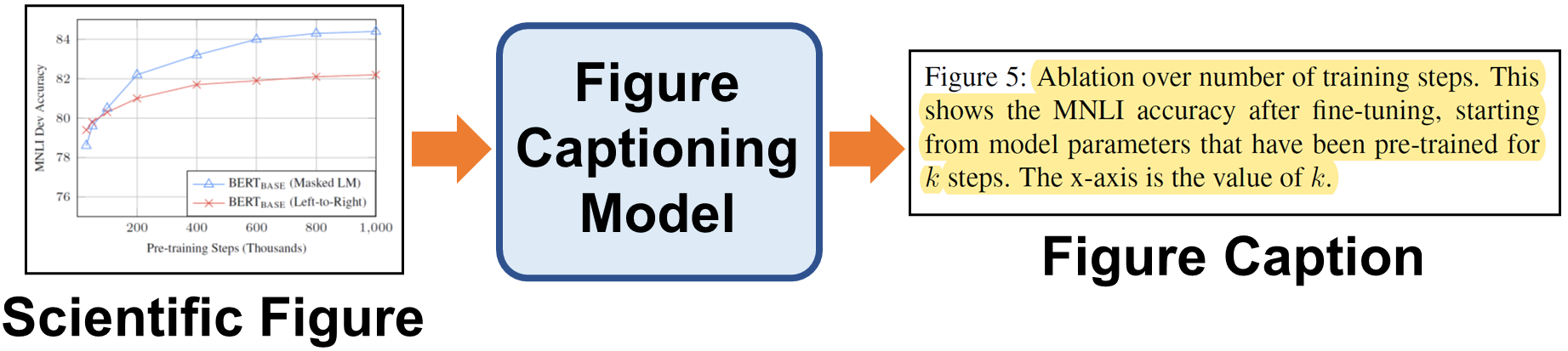}
%\vspace{-1.5pc}
\caption{The figure captioning model takes a scientific figure ({\em e.g.}, a graph plot) as input and generate captions that describes the figure.}
\label{fig:overview}
%\vspace{-1pc}
\end{figure}

Researchers use figures to explain complex concepts or show critical results.
In scholarly articles, figure captions are critical to get the message across effectively.
Ones that are too generic ({\em e.g.}, ``Results of Experiment A.'') or poorly written ({\em e.g.}, ``Relations between X and Y.'') represent missed opportunities to explain scientific narratives to readers.
Unfortunately, such low-quality captions still occur in published scientific articles.
%To address this, we will develop automated figure captioning models that generate high-quality captions for figures and charts in scientific papers (Figure~\ref{fig:overview}). Our aim is two-fold.
This paper aims to develop automatic figure-captioning models that generate high-quality captions for figures and charts in scientific papers (Figure~\ref{fig:overview}).
%Our aim is two-fold.
%Such models take a scientific figure ({\em e.g.}, a graph plot) as input and generate a caption that describes the figure.

%edited by MB
Our motivation is two-fold.
First, we aim to help researchers write better captions for the figures and charts in their papers.
Automatic caption models trained on informative, high-quality captions can suggest better captions.
Second, the proposed technology can make scientific charts and figures more accessible to blind or visually impaired readers.
Researchers have developed technologies to assist the blind to navigate graphical content, such as data visualization charts~\cite{swaminathan2014supporting}, 
printed physical maps~\cite{swaminathan2016linespace}, 
3D chemical diagrams~\cite{bernareggi2019mugraph}, 
and images on social media~\cite{wu2017automatic,salisbury2017toward}.
However, only a few prior works focused on scientific figures.
An image-captioning model specialized for scientific figures can improve the narration of scientific articles for the blind even when the original caption is unhelpful.

%edited
To this end, we introduce \textbf{\dataset}, a large-scale image-captioning dataset that contains real-world scientific figures and captions.
\dataset was constructed using computer science papers collected and released by arXiv.
With pre-processing complete -- including figure-type classification, sub-figure identification, text normalization, and caption text selection -- 
\dataset contained more than two million figures extracted from over 290,000 papers.
%\kenneth{We might want to be more specific here. What exactly are we going to release?}\edward{Didn't we plan to release the single graphplot figure? (all figures and preprocessed data)}\kenneth{But we didn't plan to release all 2M figures. We will not release the unprocessed or removed ones.}\edward{what I mean is release single graphplot figure and captions}
We then established baseline models that caption graph plots, the dominant (19.2\%) figure type.
The experimental results 
%We also established several image-captioning baselines, 
showed both exciting opportunities and steep challenges of generating captions for scientific figures.

\section{Related Work}
%The work is related to 
%{\em (i)} applications of automatic image captioning and
%{\em (ii)} technical supports for visual storytelling.

%and
%{\em (iii)} storytelling using images.

\begin{comment}

\paragraph{Applications of Image Captioning.}
Image captioning has been used to generate alternative text for images on the Internet, helping blind users navigate web pages.
Macleod {\em et al.} studied the blind's reaction to machine-generated image captions in the context of navigating online social media platforms~\cite{macleod2017understanding}.
They found that the blind tend to over-trust AI systems, even when the generated captions do not make sense.
They suggested that automatic intelligent systems should better communicate their uncertainty to users.
Researchers have developed systems to improve and polish machine-generated captions to help the blind, for example, using real-time crowdsourcing~\cite{salisbury2017toward} or reverse image search~\cite{guinness2018caption}.
Another common application is to support personal sharing.
For example, AutoCaption is a mobile application that performs a set of computer vision tasks, including face recognition,
landmark recognition, and scene recognition on photos. It then creates personalized image captions that can be shared on the social media~\cite{ramnath2014autocaption}.
Zhao {\em et al.} also developed a system that automatically captions photos for the blind before they share pictures to Facebook~\cite{zhao2017effect}.

\end{comment}

%edited MB
One of the few prior works attempting to caption scientific figures was by Chen {\em et al.}~\shortcite{Chen:2019:NCG:3341162.3345601,chen2019figure,chen2020figure}.
%\edward{\cite{chen2019figure}\cite{chen2020figure}}.
%Chen {\em et al. to caption scientific figures~\shortcite{Chen:2019:NCG:3341162.3345601}.
They created FigCAP, a caption-figure pair corpus where the figures are synthesized, and used an LSTM model with an attention mechanism to produce captions.
%This work is inspiring and relevant, but it did not	use real-world figure-caption pairs. 
%\kenneth{Move text from rebuttal or Adobe Proposal to explain WHY}
%Chen explained that they used synthetic data because the realistic dataset was relatively small.
%In this paper, we take advantage of the large-scale paper dataset collected and release by arXiv to develop our model.
%Furthermore, 
FigCAP was built on research that aimed to analyze figure content automatically,
%in figure understanding, which attempted to analyze figure content automatically.
%is the extraction of the plotted data and its association
%with the legend entries
%This 
including Figure-Seer~\cite{siegel2016figureseer}, 
FigureQA~\cite{kahou2017figureqa}, and
DVQA~\cite{kafle2018dvqa}.
%The figures in both 
DVQA and FigureQA were both made using synthetic figures;
FigureSeer contained over 60,000 figures across seven figure types extracted from research papers.
Meanwhile, Qian {\em et al.}~\shortcite{qian2020formative} proposed a set of ``caption units'' (such as Title, Label Name, Min/Max, etc.) that are important to include in a caption of scientific figures; they created a model, FigJAM, to produce such units~\cite{qian2021generating}.
%Chart-to-Text: Generating Natural Language Descriptions for Charts by Adapting the Transformer Model
%An overview of the proposed approach for chart summarization using a transformer-based model. 
%The model takes the data table and some chart metadata as input (on the left) and generates a summary containing data
%variables that refer values within the data table
%Neural Data-Driven Captioning of Time-Series Line Charts
Also relevant is the ``data-to-caption'' work, which takes a chart's source data table and metadata as input to generate a caption~\cite{obeid2020chart,spreafico2020neural}.
These models generate captions based on data tables, not the figures.
%well-known “six hats” method, which asks
%people to wear metaphorical hats representing different thinking perspectives [10]. Teevan et al. proposed to use the six
%hats schema to assign different thinking roles to the authors
%themselves in order to promote self-reflection from different
%angles [38]. Chou et al. showed that perspective-taking can

%that achieves this goal through utilizing metadata
%information and a joint static and dynamic dictionary

%Our study results show that real-world captions usually consist of a finite set of caption units and that automatic figure
%captioning should be formulated as a multi-stage task.

%\kenneth{WHY it's different?}
%\kenneth{Add Xin's CHI LBW and WWW papers.} \cite{qian2021generating}\cite{qian2020formative}\edward{added}

\paragraph{Differences Between Synthetic and Real-World Captions.}
%Using human-written captions to develop captioning models will allow us to understand how scientists write captions in papers.
Most prior work has tried to generate captions for scientific figures using synthetic images and texts~\cite{Chen:2019:NCG:3341162.3345601,chen2019figure,chen2020figure,kahou2017figureqa}.
%\edward{\cite{chen2019figure}\cite{chen2020figure}}
%Real-world captions are better because they encode the realistic languages used in scientific captions.
However, synthetic captions tend to be generic and describe features without conveying higher-level insights,
for example,
\textit{``This is a line plot. It contains 6 categories. Dark Magenta has the lowest value. Lawn Green has the highest value.''} (example from FigCAP.)
%\edward{\cite{chen2019figure}\cite{chen2020figure}}.\kenneth{Is this example indeed from this dataset? What's the dataset name?}\edward{real example from FigCAP}
Human-written captions, on the other hand, tend to highlight the meaningful parts of the figure and bring more context, for example: 
\textit{``Train loss curve with respect to optimization steps. With prior coarse-tuning on NLI data, convergence becomes much faster and easier.''}
%Example of learning curve for unigram BLEU score. Red line is the exact BLEU score obtained by... Purple dotted line is... Crosses denote expected BLEU score calculated by averaging samples from distri- bution px.
[example from~\cite{jin2020mmm}].

\section{Constructing \dataset Dataset}
\label{sec:dataset-construct}

%\kenneth{The size of data in each step is important. Make sure you keep track of all of them.}\edward{correct}

%edited by MB
%Most prior work used synthesized data to train figure-captioning models.
%Real-world figures and captions, on the other hand, are much richer and also much nosier.\kenneth{noisier is not so easy to understand.}\edward{contain more diverse and details information}
This section describes the process that massages real-world figure-caption data into an appropriate easy-to-use format for the NLP community.
This data-processing procedure was developed iteratively and empirically.

%edited by MB
\paragraph{Step 1: Data Acquisition and Pre-processing.}
Data acquisition is a fundamental challenge for constructing a public scientific figure-caption dataset.
Although there is a vast number of scientific papers, they are not all easy to access.
\dataset is based on the arXiv dataset~\cite{clement2019arxiv}.\footnote{arXiv Dataset on Kaggle: \url{https://www.kaggle.com/Cornell-University/arxiv}}
%\cy{Is it from this paper ``ON THE USE OF ARXIV AS A DATASET''?}\edward{yes, and Kaggle found it}\cy{Cite it as well then}\edward{\cite{clement2019arxiv}}
The arXiv dataset is licensed under CC-0, which grants remake and republish rights.
It contains a repository of 1.7 million articles with relevant features, such as article titles, authors, categories, abstracts, full-text PDFs, and more.

%edited by MB
We first downloaded all the scholarly articles from the arXiv dataset and froze the date on Dec 22, 2020 (a total of 1,921,287 papers). 
\dataset does not include any papers published after this date.
We further narrowed our dataset to papers published between 2010 and 2020 in computer science (cs.) and machine learning (stat.ML) topics, which numbered 295,028 papers.
%\kenneth{make sure the number is correct}\edward{correct}
We did not use these papers' ``source files,'' which might contain the original LaTeX and figure files. Not all papers come with source files; some source files have complex dependencies that are hard to parse.
%\cy{Change the following paragraph?}
%Note that we did not use the papers' source files, which might include the original LaTeX files and figure files.
%We did not use these source files because not all papers have source files. 
%Some paper's source files even have complex dependencies that are not easy to parse.
%\cy{Although source files might contain the original LaTeX and figure files,
%not all papers come with the source files and these source files might have complex dependencies that are hard to parse.}

%edited by MB
\paragraph{Step 2: Figure-Caption Pair Extraction.}
We then used PDFFigures 2.0~\cite{clark2016pdffigures} to extract the figures from papers in our paper collection.
PDFFigures 2.0 is a Scala-based tool created to extract figures, captions, tables, and section titles from scholarly documents, with a focus on
% documents from the domain of computer science.
the computer science domain.
%edited by MB
In addition to the figures' images and captions, the tool also extracted all
%the exact coordinates of those components\kenneth{What components exactly?} \edward{they give every figure and text coordinates of region in the paper}and 
the text snippets inside the figures, such as legends, X-Y labels, and titles.
%\kenneth{What exactly is the text snippets? Title? Legend?}\edward{any text inside the figure, legend, x-y labels, title if in figure}
%Potentially, 
The extracted information can be used to boost the performance of image-captioning models.
%\kenneth{We can remove this to save space if needed.}
This step resulted in 295,028 papers and 2,170,719 figures.

%edited by MB
\paragraph{Step 3: Figure Type Classification.}
Given the high diversity in the figure types included in scientific articles, we did not aim to create a single captioning model for all types of figures.
Instead, we aimed to create captioning models specialized for one particular figure type.
%From a writing support perspective, the author of the paper  
%\kenneth{Add some writer scenarios maybe?}
We used an automatic figure type classifier~\cite{siegel2016figureseer} to classify	figure type in \dataset.
This pre-trained classifier	can identify seven types of figures: 
graph plots,
flowcharts (also called node diagrams), 
equations (also called algorithms), 
bar plots, 
scatter plots, 
tables, 
and ``other.''
Its reported accuracy is 86\% over 60,000 samples~\cite{siegel2016figureseer}.

%edited by MB
According to the classifier's prediction, out of 2,170,719 figures, 
19.2\% (416,804) are graph plots,
23.6\% (511,984) are tables,\footnote{In this work, tables are not considered to be figures due to drastically different visual features and contents.}
5.9\% (127,197) are	equations (including algorithms and pseudo codes),
8.5\% (185,398) are flowcharts, 
2.0\% (44,052) are scatter plots, 
4.7\% (101,146) are bar charts, and	
36.1\% (784,138) are ``other.''
In \dataset, we only focus on graph plots, which have the highest classification performance~\cite{siegel2016figureseer} and are also the most common figure type.
%()
% Due to drastically different visual features, content, and uses, we did not consider tables to be figures.
% \cy{Do you ``only'' consider ``graph plots''? or only ``tables'' are removed?} \edward{only consider graph plots}

%edited by MB
\paragraph{Step 4: Removing Figures with Subfigures.}
%Real-world data is rich but noisy; the quality, length, and amount of information varies significantly among captions in published papers.
Many scientific figures contain subfigures.
For example, in our pilot study	(Section~\ref{sec:2000-figure-study}), 35.72\% of overall scientific figures had subfigures.
\dataset focuses on generating captions for single figures, so we removed figures with %more than one 
subfigures from the dataset.
We first used handcrafted rules to identify captions that explicitly mention or refer to subfigures [for
example, \texttt{(a)}, \texttt{a)}, \texttt{(b)}, \texttt{b)}, \texttt{(1)}, \texttt{1)}, \texttt{(2)}, \texttt{2)} ... etc.]. 
%This rule-based approach yielded high classification precision but low recall.
%(See Section~\ref{sec:2000-figure-study}).
%\kenneth{Are these numbers based on the 2000 images you labeled?}\edward{yeap}
%To increase the recall, 
Furthermore, we also used FigureSeparator~\cite{tsutsui2017data} to filter figures with subfigures out of our collection.
FigureSeparator is a CNN-based model 
that separates compound figures in the ImageCLEF Medical dataset with 85.9\% accuracy.

%that separates compound figures with 85.9\% accuracy on the ImageCLEF Medical dataset. 
Of 416,804 graph plots identified in Step 3, the rule-based approach yielded
%first removed images with subfigures and kept
352,719 graph plots, and the FigureSeparator further narrowed the collection down to 133,543 figures.
%Namely,
An estimated 32.04\% of the graph plots did not have subfigures.
%were passed to the next step.
%416,804 啥都沒做的graphplot
%rule-based之後3522719
%rule-based+separator之後 133543
%416,804 * 32.04% = 133543
%416,804 * 67.96% = 283261
%Of 133,543 graph plots identified in the previous step, 67.96\% contained subfigures, which we removed from our dataset.
%\kenneth{CHECK NUMBER. This number is still incorrect. ``Of 133,543 graph plots identified in the previous step'', but in Step 3 we said ``416,804'' graph plot}
%\edward{32.04\% are single figure, 67.96\% are multiple figures}
%\kenneth{Q for Ed: If this is the case, WHY in Section~\ref{sec:2000-figure-study} only 35.72\% of figures DID HAVE subfigures? Because Graph Plot has more subfig??? }

%edited by MB
\paragraph{Step 5: Text Normalization.}
%In \dataset, 
We used NLTK~\cite{Loper02nltk:the} for tokenization and converted all the text to lowercase.
We also removed the figure numbers, such as ``Figure 1:'' or ``Fig. 1:'', and only kept the main caption text.
%and then applied 
The following two text normalization strategies were then applied:

%\vspace{-.4pc}

\begin{itemize}

%edited by MB
\item \textbf{Basic Normalization:}
We replaced all the numbers ({\em e.g.}, \texttt{0}, \texttt{-0.2}, \texttt{3.44\%}, \texttt{1,000,000}) with \texttt{[NUM]}.
%\kenneth{Did I miss any thing? Negative sign, percentage, interger...}
%We also removed all instances of the percentage sign (\%).
%\kenneth{What does this sentence mean? removed all instances of the percentage?}
%\kenneth{Q for Ed: Did you do the tokenization and lowercase for the first raw in the experiments? I thought you at least need to do lowercase?...}\edward{yes, I also do naive normalization that tokenize, lowercase and remove the initial title (e.g. Figure x./fig. x.)}

%\vspace{-.5pc}

%edited by MB
\item \textbf{Advanced Normalization:}
%We further performed advanced normalization operations.
We created regular expressions
%We adapted the regular expressions used in ARQMat~\cite{mansouri2020finding}, which was created to recognize equations in LaTeX codes, 
%which is powered by a rich set of regular expressions, 
to identify equations in captions and replaced them with \texttt{[EQUATION]}.
%\kenneth{Do we really need to cite ARQMat? Ed, if we just use some heuristics to find equations, %we might not need to cite ARQMat.}\edward{you may remove it}
%\kenneth{How exactly did you adapt their regular expressions? Example?}\edward{simple way, if [words] = [NUM]}
We also replaced all the text spans enclosed by any types of bracket pairs, including \texttt{\{\}}, \texttt{[]}, and \texttt{()}, with \texttt{[BRACKET]}.

%\kenneth{ADD explanation about it can be improved.} \kenneth{Add examples why it worked?} \edward{Since the regular expression method used in \cite{mansouri2020finding} is used in latex format contains special symbols which is easier and more accuarate to recognize the range of math equation compare to pure text. Other methods could be either train a model to label math equation in text or use image recognition of math equation purposedd by \cite{mali2020scanssd}} 
%\kenneth{Give me (1) some example regular expressions and (2) some before/after captions here.}

\end{itemize}

%\vspace{-.7pc}

%edited by MB
\paragraph{Step 6: Target Caption Text Selection.}
\dataset provides three different data collections, each sampled using different strategies:

%\kenneth{We also calculated the percentage of one-sentence captions and the average length of a caption:}\edward{do you want avg sentences/tokens}
%[Data Cleaning Method: % of caption w/ one sentence, Avg. #sentence per caption]
%Original: 59.42%, 1.87
%Rule-Based: 63.98%, 1.70
%Rule-Based+FigureSeparator: 70.47%, 1.51

%\vspace{-.5pc}

\begin{itemize}

%edited by MB
\item \textbf{First Sentence (133,543 Figures):} 
This collection includes all the figures.
For each figure included, this collection only includes the first sentence of the caption.
%
%\vspace{-.5pc}

%edited by MB
\item \textbf{Single-Sentence Caption (94,110 Figures):} 
This collection includes the complete caption of only the figures with a one-sentence caption.
%Namely,  
Of the graph plots, 70.47\% had a one-sentence caption.
%Note that this collection is a subset of the First Sentence collection.
%\kenneth{ADD NUMBER HERE, 94110/133543 = 70.47 maybe?}
%\vspace{-.5pc}

%edited by MB
\item \textbf{Caption with No More than 100 Words (131,319 Figures):} 
This collection includes the complete caption of only the figures %with a one-sentence caption. 
%This collection includes only the figures 
whose captions contained no more than one hundred tokens (punctuation marks included).
In this collection, a caption contains 1.66 sentences on average (SD=1.07).
%\kenneth{ADD NUMBER!!!}\edward{done}
%\kenneth{Does 100 tokens include punctuation marks (?!.,)?} \edward{yes}
%'s first sentence had no more than 100 to- kens.
%For each figure included, this collection includes the maximum number of sentences (counted from the beginning of the caption) be- fore exceeding 100 tokens.

\end{itemize}

%\vspace{-.5pc}

On average, with advanced normalization (Step 4),
a sentence in the ``First Sentence'' collection contains 23.19 tokens (SD=20.86);
a sentence in the ``Single-Sentence Caption'' collection contains 14.05 tokens (SD=8.15);
and a sentence in the ``Caption with No More Than 100 Words'' collection contains 22.04 tokens (SD=17.44).
%\kenneth{Oh, so the ``First Sentence'' collection is NOT shorter than ``Caption with No More Than 100 Words'' collection? Because some of the first sentences are super long?}\edward{I think is longer? In the advanced setting, yes. There indeed some cases first sentence super long, I can find some examples if needed.}
%\kenneth{Typo, I missed the ``NOT'': Is the ``First Sentence'' collection NOT shorter???}\edward{Let me parse to see}
%\kenneth{ADD NUMBERS HERE!!!}\edward{done}
%\kenneth{TODO: Add SD}\edward{done}

%Training, validation, and test sets 
Note that we first created the 80/10/10 train/val/test data split for the entire corpus and then proceeded with the caption selection step.
This procedure ensured that we used the identical set of figures to construct each collection's test set; the
same applied to their training and validation sets.
%For example, if the First Sentence collection's test set contains a figure with a ten-word single-sentence caption,
%the same figure occurs in the test sets of the other two collections, too. \kenneth{This FOR EXAMPLE feels quite extra...}

%For example:

%+Figure X’s caption has 1 sentence. Figure X is included in the test set of (a), (b), and (c). (a) and (b) uses the whole one sentence; (c) uses the first 100 words.
%+Figure Y’s caption has 2 sentences. Figure Y is included in the training set of (a) and (c). (a) uses the first sentence; (c) uses the first 100 words. (b) doesn’t select this figure because it only selects captions with a single sentence.

%So, if the Text Data is different, the figures in the test/dev/train splits are roughly the same, but the caption texts are different. 

%Despite all the data processing and filtering strategies applied in the previous steps, the remaining figures still had quite a diverse range of caption types. 

%\input{tech-eval-table}

% 單一分類器的情況
% TN: 真多張 = (分類器說是多張, 實際上也是多張)  
% FN: 假多張 = (分類器說是多張, 實際上卻是單張)  
% TP: 真單張 = (分類器說是單張, 實際上也是單張)  
% FP: 假單張 = (分類器說是單張, 實際上卻是多張)

% 兩個分類器合起來的情況
% TN: 真多張 = (至少有一個分類器說是多張, 實際上也是多張)
% FN: 假多張 = (至少有一個分類器說是多張, 實際上卻是單張)  
% TP: 真單張 = (兩個分類器都說是單張, 實際上也是單張)  
% FP: 假單張 = (兩個分類器都說是單張, 實際上卻是多張)

% \usepackage{booktabs}
\begin{table}[t]
\footnotesize
\centering
\begin{tabular}{@{}lrrrr@{}}
\toprule
\rowcolor[HTML]{EFEFEF} 
\multicolumn{5}{c}{\cellcolor[HTML]{EFEFEF}\textbf{Figure Type Classification} (Class = Graph Plot)}  \\ \midrule
\rowcolor[HTML]{FFFFFF} 
\textbf{Approach}          & \textbf{P} & \textbf{R} & \textbf{F} & \textbf{Acc} \\ \midrule
\cite{siegel2016figureseer}              & .90        & .83        & .87        & .95          \\ \midrule
\rowcolor[HTML]{EFEFEF} 
\multicolumn{5}{c}{\cellcolor[HTML]{EFEFEF}\begin{tabular}[c]{@{}c@{}}\textbf{Non-Subfigure Figure Classification}\\ (For figures labeled as graph plots in Step 3.)\end{tabular}} \\ \midrule
\rowcolor[HTML]{FFFFFF} 
\textbf{Approach}          & \textbf{P} & \textbf{R} & \textbf{F} & \textbf{Acc} \\ \midrule
Rule-Based                 & .54        & .95        & .69        & .59          \\ \midrule
FigureSeparator            & .98        & .66       & .79        & .83          \\ \midrule
Rule-Based+FigureSeparator & .98        & .62        & .76        & .81          \\ \bottomrule
\end{tabular}
\caption{The tools used to construct \dataset evaluated on 1,926 labeled images.
For figure type classification, the overall performance over graph plots was reliable.
Regarding identifying the graph plots (as labeled automatically in Step 3) that do not contain subfigures, FigureSeparator achieved an exceptionally high precision.}
\label{tab:tech-eval}
\end{table}

\subsection{Data Analysis and Quality Measurement}
\label{sec:2000-figure-study}

%edited by MB
To evaluate the quality of our data cleaning and processing pipeline, we randomly sampled 2,000 figures from the original arXiv dataset, and one author manually labelled each figure's figure type and whether it contained subfigures (Yes/No).\footnote{To validate the label quality, we had three graduate students label 100 figures, respectively.
On average, they %three annotators 
agreed with 97\% of our subfigure labels.
%\kenneth{TODO: Use Ed's subfig (Y/N) labels as gold-standard label and calculate each of their accuracy (?/100) and then average the accuracy.}\edward{done}
For the figures without subfigures, they agreed with our figure type labels 82.17\% of the time.
%\kenneth{TODO: Similarly, calculate their average accuracy}\edward{done}
For the figures with subfigures, they agreed with at least one of our type labels 86.56\% of the time.
%\kenneth{TODO: You can probably use hand to count numbers here. We only have like around 30 figures with subfigures?}\edward{done}
}
Of these 2,000 figures, 1,926 figures had no extraction errors, and were included in our follow-up calculation.
%\kenneth{Do we have anything like Kappa here?}\edward{what does that mean?}\kenneth{Did you compare Hua/CY/Chacha's labels with yours, or with each other, in any ways? Could we say something here?}
%\kenneth{How exactly you label the data? Each figure was annotated by one or two person?}\edward{I create interface and annotated 2000 samples myself}\kenneth{Wait I thought CY Chacha and Hua helped???}\edward{they just do cross validation for 100 examples, nobody would like to annotate 2000 sample except me}
%\kenneth{Explicitly talk about multi-type subfigures. --> We annotate multi class since this is more realistic.}
As for types, 20.35\% of the figures were graph plots, 4.1\% were bar charts, and 3.11\% were scatter plots.\footnote{A figure might contain subfigures of different types ({\em e.g.}, a bar chart accompanied by a graph plot.) For each figure, we took a multi-class labeling strategy that exhaustively labels all distinct types of its subfigures.}
%\kenneth{This is the FIGURE type, correct?}\edward{correct}
%\kenneth{We might want to explain complex relations between subfigs and types.}
In terms of subfigures, 237 out of 1,926 figures (35.72\%) contained subfigures:
33.14\% of these figures contained graph plots as subfigures,
5.81\% contained bar charts, and 
6.83\% contained scatter plots.

%\kenneth{Since we used multi-class labels, the word ``contained'' is probably more accurate than `` were''?}\edward{okay}
%\kenneth{TODO: Do we consider subfig here?}\edward{what does that mean?}

%edited by MB

We used these 1,926 labeled images to evaluate the tools we employed in constructing \dataset.
Table~\ref{tab:tech-eval} shows the results.
For the figure type classification, the overall performance over graph plots were reliable.
Regarding identifying the graph plots (as labeled automatically in Step 3) that do not contain subfigures, FigureSeparator had an exceptionally high precision.

\section{Experimental Results}
%\input{table-new.tex}

% \input{train_val_test_size}

% \usepackage{booktabs}
% \usepackage{multirow}
\begin{table}[t]
\footnotesize
\centering
\begin{tabular}{@{}ccccccc@{}}
\toprule
\multicolumn{7}{c}{\cellcolor[HTML]{EFEFEF}\textbf{First Sentence}} \\ \midrule
\multicolumn{2}{c}{\textbf{Subfig Filter}} &
  \multicolumn{2}{c}{\textbf{Norm.}} &
   &
   &
   \\ \cmidrule(r){1-4}
\textbf{Rule} &
  \textbf{FigSep} &
  \textbf{B.} &
  \textbf{A.} &
  \multirow{-2}{*}{\textbf{\#Fig.}} &
  \multirow{-2}{*}{\textbf{\begin{tabular}[c]{@{}c@{}}Vocab\\ Size\end{tabular}}} &
  \multirow{-2}{*}{\textbf{BLEU-4}} \\ \midrule
 &
   &
   &
   &
  416,804 &
  30,776 &
  .0259 \\ \midrule
\cmark &
   &
  \cmark &
   &
  352,719 &
  24,355 &
  .0236 \\ \midrule
\cmark &
  \cmark &
  \cmark &
   &
   &
  12,666 &
  .0224 \\ \cmidrule(r){1-4} \cmidrule(l){6-7} 
\cmark &
  \cmark &
  \cmark &
  \cmark &
  \multirow{-2}{*}{133,543} &
  11,946 &
  .0219 \\ \midrule
\multicolumn{7}{c}{\cellcolor[HTML]{EFEFEF}\textbf{Single-Sentence Caption Only}} \\ \midrule
\multicolumn{2}{c}{\textbf{Subfig Filter}} &
  \multicolumn{2}{c}{\textbf{Norm.}} &
   &
   &
   \\ \cmidrule(r){1-4}
\textbf{Rule} &
  \textbf{FigSep} &
  \textbf{B.} &
  \textbf{A.} &
  \multirow{-2}{*}{\textbf{\#Fig.}} &
  \multirow{-2}{*}{\textbf{\begin{tabular}[c]{@{}c@{}}Vocab\\ Size\end{tabular}}} &
  \multirow{-2}{*}{\textbf{BLEU-4}} \\ \midrule
 &
   &
   &
   &
  247,649 &
  21,765 &
  .0291 \\ \midrule
\cmark &
   &
  \cmark &
   &
  218,655 &
  17,685 &
  .0228 \\ \midrule
\cmark &
  \cmark &
  \cmark &
   &
   &
  9,760 &
  .0234 \\ \cmidrule(r){1-4} \cmidrule(l){6-7} 
\cmark &
  \cmark &
  \cmark &
  \cmark &
  \multirow{-2}{*}{92,021} &
  9,232 &
  .0207 \\ \midrule
\multicolumn{7}{c}{\cellcolor[HTML]{EFEFEF}\textbf{Caption with \textless{}= 100 Words}} \\ \midrule
\multicolumn{2}{c}{\textbf{Subfig Filter}} &
  \multicolumn{2}{c}{\textbf{Norm.}} &
   &
   &
   \\ \cmidrule(r){1-4}
\textbf{Rule} &
  \textbf{FigSep} &
  \textbf{B.} &
  \textbf{A.} &
  \multirow{-2}{*}{\textbf{\#Fig.}} &
  \multirow{-2}{*}{\textbf{\begin{tabular}[c]{@{}c@{}}Vocab\\ Size\end{tabular}}} &
  \multirow{-2}{*}{\textbf{BLEU-4}} \\ \midrule
 &
   &
   &
   &
  395,024 &
  37,885 &
  .0231 \\ \midrule
\cmark &
   &
  \cmark &
   &
  341,350 &
  30,316 &
  .0098 \\ \midrule
\cmark &
  \cmark &
  \cmark &
   &
   &
  15,642 &
  .0173 \\ \cmidrule(r){1-4} \cmidrule(l){6-7} 
\cmark &
  \cmark &
  \cmark &
  \cmark &
  \multirow{-2}{*}{132,120} &
  14,974 &
  .0172 \\ \bottomrule
\end{tabular}
%\vspace{-.5pc}
\caption{The baseline model's performance on \dataset, using Vision-Only features.
Models trained on the Single-Sentence Caption collection performed the best.
%; normalizing caption improves the model's performance.
%; the text information (T) extracted from figures does not help the models. 
The low BLEU-4 scores indicate that more research is needed to reliably generate captions for scientific figures. 
%\kenneth{The vocab size is BEFORE pruning, right?}
(The vocabulary sizes were calculated after dropping words with a frequency below 5.)
%\kenneth{TODO: Move all the numbers to the new table}
}
%\vspace{-1.2pc}
\label{tab:text-only-result}
\end{table}

% \usepackage{booktabs}
\begin{table}[t]
\footnotesize
\centering
\begin{tabular}{@{}llr@{}}
\toprule
\textbf{Data Collection}                          & \textbf{Feature} & \textbf{BLEU-4} \\ \midrule
\multirow{3}{*}{First Sentence}                   & Vision Only      & .0219           \\ \cmidrule(l){2-3} 
                                                  & Vision+Text      & .0205           \\ \cmidrule(l){2-3} 
                                                  & Text Only        & .0213           \\ \midrule
\multirow{3}{*}{Single-Sent Caption}              & Vision Only      & .0207           \\ \cmidrule(l){2-3} 
                                                  & Vision+Text      & .0202           \\ \cmidrule(l){2-3} 
                                                  & Text Only        & .0212           \\ \midrule
\multirow{3}{*}{Caption w/ \textless{}=100 words} & Vision Only      & .0172           \\ \cmidrule(l){2-3} 
                                                  & Vision+Text      & .0168           \\ \cmidrule(l){2-3} 
                                                  & Text Only        & .0165           \\ \bottomrule
\end{tabular}
%\vspace{-.5pc}
\caption{The experimental results of models using Vision-Only, Text-Only, and Vision+Text features. Vision-Only and Text-Only features yielded similar performance. (All the subfigure-filtering and text-normalization steps were applied.)}
% The text extracted from figures did not help the models. 
% (All the subfigure-filtering and text-normalization steps were applied.)
%\edward{this set use the last one setting}}
\label{tab:text-and-vision-result}
%\cy{Do we have ``Vision Only'' and ``Text Only'' results for Single-Sentence and Caption with \textless{}= 100 Words?}
%\kenneth{This table needs to have 9 rows.}
%\vspace{-1pc}
\end{table}

\begin{figure*}[t]
\centering
\includegraphics[width=0.91\textwidth]{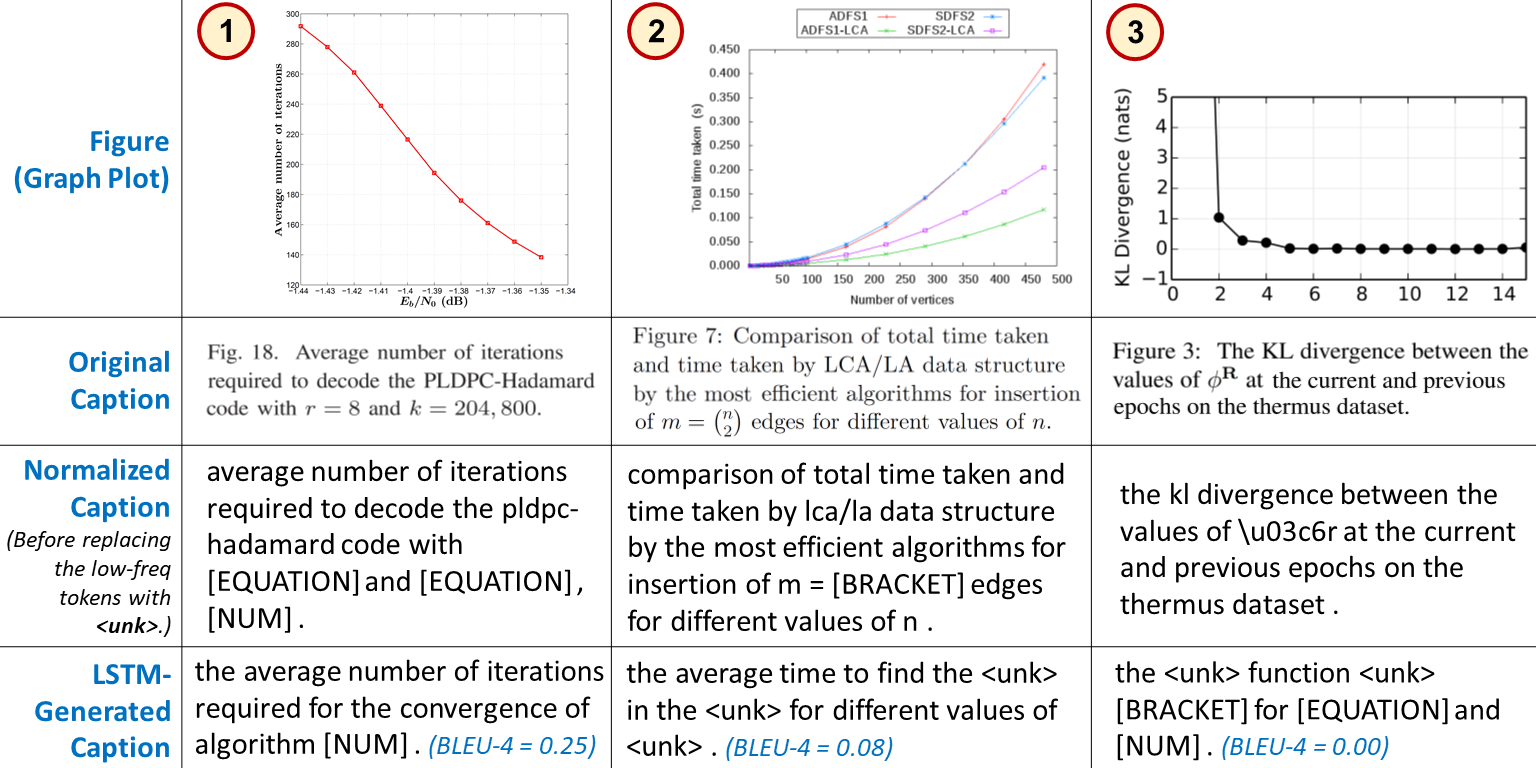}
%\vspace{-.5pc}
\caption{Example outputs of the baseline models trained and tested on the Single-Sentence Caption Only collection.
Intensive research will be needed to create models that can caption scientific figures reliably. 
[Figure sources: (1)~\cite{zhang2020protograph}, (2)~\cite{baswana2017incremental}, and (3)~\cite{brubaker2015building}.]}
\label{fig:example}
%\vspace{-.7pc}
\end{figure*}

%edited by MB
To examine the feasibility and challenges of creating an image-captioning model for scientific figures, we established several baselines and tested them using \dataset.
The caption quality was %automatically 
measured by BLEU-4~\cite{papineni2002bleu}, using the test set of the corresponding data collection as a reference.
Figure~\ref{fig:example} shows some example outputs.

%edited by MB
\paragraph{Baseline Model.}
\cy{I also rewrite the model description. Please check. @Edward please check if there is anything wrong.}
We used a classical image-captioning model, CNN+LSTM architecture, as our baseline~\cite{xu2015show}.
The pre-trained ResNet-101~\cite{he2016deep} was used as the image encoder to represent a figure as a 2048-dimension vector.
This image vector was then fed into a dense layer to fit the dimension of the word-embedding and the LSTM decoder where 
the word-embedding and LSTM hidden layer size were all 512.
A global attention mechanism was added to the LSTM decoder to better model the context~\cite{luong2015effective}.
The LSTM decoder took the image vector as the initial state and generate captions.

We designed three variations of the baseline models, Vision-only, Vision+Text, and Text-only.
The text information was the titles, legends, and X-Y labels extracted from the figures (Step 2 in Section~\ref{sec:dataset-construct}).
Another LSTM was used as a text encoder to encode text information into a vector.
For the Vision+Text variation, we concatenated the image vector and the text vector together 
and fed it into the LSTM decoder for caption generation.
The Text-only variation only took the text vector as the feature for the LSTM decoder.

%edited MB
\paragraph{Experimental Setups.}
We trained the baseline models using an 80/10/10 train/val/test data split.
The models were trained by minimizing a cross-entropy loss with a doubly stochastic regularization~\cite{xu2015show}
using Adam~\cite{kingma2014adam}.
The weights of the pretrained ResNet-101 image encoder were partially frozen so that only 
convolutional blocks 2 through 4 were fine-tuned throughout the training process~\cite{yosinski2014transferable}.
We empirically set the hyper-parameters by observing the performance gain on the validation set.
Hyper-parameters ended up being used were a dropout rate of 0.5; a batch size of 16/32; 
a learning rate of 4e-4 with a decay factor of 0.8 when there was no improvement for 8 epochs.
The models were trained until there was no improvement for 20 epochs.
We kept the model with the highest BLEU-4 score on the validation set for testing.

%We implement the following models with Pytorch, and conduct experiments on a single nVidia GeForce RTX 2080 GPU. 

%}\edward{do we need to give our model details?}

\paragraph{Results.}
%\kenneth{!!!This part needs to rewrite to fit in the new table.}
%Note that 
We trained the %baseline 
models on each data collection with varying levels of data filtering and text normalization.
Table~\ref{tab:text-only-result} shows the results.
%As mentioned in the Step 3 of Section~\ref{sec:dataset-construct}, FigureSeperator %different combined settings of data.
%As defined previously, 
%Text Data represented different targeted captions; 
%Cleaning Method represented different subfigure classification; 
%Text Normalization represented post-processing captions;
%Feature represented the information we used as training data; 
%and the last column displays the performance of generated captions.
%edited by MB
%\paragraph{Evaluations.}
%We report baseline models' performances in Table~\ref{tab:text-only-result}.
Among the three data collections, the models trained on the single-sentence captions performed the best. %\kenneth{This need to update later...}
This might be because the Single-Sentence Caption collection, which is a subset of the First Sentence collection, 
had the smallest vocabulary size.
%This might be because the first sentence of a caption is often the topic sentence, and thus has a lower lexical diversity.
%\cy{Is this true? Do we have numbers? As ``Single-sentence'' is a subset of ``First-sentence'', I feel it might have a even lower lexical diversity? I am not really sure about the relationship between ``topic sentence'', ``lower lexical diversity'', and ``performance''.}
%\kenneth{You're right, MAYBE it's just data size is lareger?}

%edited by MB
\paragraph{Effects of Text Normalization.}
%We performed different degrees of text normalization in \dataset to see whether text normalization could increase the quality of the generated captions. 
Our experiments did not show the clear benefits of normalizing text to the resulting BLEU-4 scores.
We will explore other methods to normalize text, for example, using advanced techniques to identify equations in text~\cite{mali2020scanssd,mansouri2020finding}.

%We adapted the regular expressions used in ARQMat~\cite{mansouri2020finding}, which was created to recognize equations in LaTeX codes, 
%which is powered by a rich set of regular expressions, 

%edited by MB
\paragraph{Effects of Text and Vision Features.}
%\kenneth{Add some text to describe new tables}
We also used Vision-Only, Text-Only, and Text+Vision features to develop models (Table~\ref{tab:text-and-vision-result}). 
%Our experiments showed that 
Vision-Only and Text-Only features yielded similar performance.
%The models performed similarly when using either Vision-Only or Text-Only features.
Furthermore, the models performed slightly worse when training on combined features.

\section{Conclusion and Future Work}
%edited by MB
This paper introduces \dataset, a large-scale image-captioning dataset that contains real-world scientific figures and captions. 
%\dataset was constructed using the papers collected and released by arXiv, containing more than two million figures were extracted from over 290,000 papers.
\dataset was constructed using more than two million images from over 290,000 papers collected and released by arXiv.
We also established	several image-captioning baselines, showing the feasibility and challenges of generating captions for scientific figures.
In the future, we will explore approaches to improve caption quality, such as taking advantage of large pre-trained language models~\cite{Beltagy2019SciBERT}, %in specific domains, 
or using information in paper's full text to boost performance.
%or learning to post-edit the generated captions~\cite{hsu-etal-2019-visual}.

%\kenneth{Say slightly more about using the paper}
%\kenneth{One of the future direction can also be post-editing. Cite our own paper.}

%------------- dead kitten ----------

\begin{comment}

% \kenneth{UPDATE NUMBERS ACCORDINGLY}
This paper introduces \dataset, a large-scale image captioning data that contains real-world scientific figures and captions.
\dataset was constructed using the papers collected and released by arXiv,
%\dataset 
containing more than two million figures were extracted from over 290,000 papers.
% We focus on graph plots which is the most dominant and clear defined category of scientific figures. We further develop neural image caption models that generate high-quality captions for those figures.
We also establish several image captioning baselines, showing the feasibility and challenges of generating captions for scientific figures.

In the future, we will explore novel approaches to improve the caption quality.
For example, taking advantage of large pre-trained language model in specific domain, or use the extra information in the paper, to boost the performance.

\end{comment}

\section*{Ethical Considerations}

%edited by MB
\paragraph{Data Licensing.}
The arXiv dataset uses the CC0 1.0 Universal (CC0 1.0) Public Domain Dedication license,\footnote{CC 1.0: \url{https://creativecommons.org/publicdomain/zero/1.0/}} which grants permission to remix, remake, annotate, and publish the data.

\paragraph{Potential Biases of Language Technologies.}
We are aware that language technologies	trained on a ``standard'' or mainstream variety of a language (in our case, English) favor the popular variety and harms people using varieties with fewer speakers.
For example, standard automatic speech recognition trained on Dutch speeches results in 10-15\% higher error rates on Flemish Dutch than on ``standard'' Dutch~\cite{feng2021quantifying}.

%---------------- dead kitten --------------

\begin{comment}

\paragraph{Potential Biases of Language Technologies.}
We are aware of that the language technologies trained on ``standard'' or mainstream variety of a language (in our case, English) would favoring the popular variety and harms the people using the varieties with fewer speakers.
For example, the standard automatic speech recognition trained on Dutch speeches results in a 10\% to 15\% higher word error rates on Flemish Dutch than that of on ``standard'' Dutch~\cite{feng2021quantifying}.

\paragraph{Data Licensing.}
The arXiv dataset uses CC0 1.0 Universal (CC0 1.0) Public Domain Dedication license,~\footnote{CC 1.0: https://creativecommons.org/publicdomain/zero/1.0/} which permits use to remix, remake, annotate, and publish the data.

\end{comment}

%Acknowledgments
\section*{Acknowledgments}
We thank Chieh-Yang Huang, Hua Shen, and Chacha Chen for helping with the data annotation.
We thank Chieh-Yang Huang for the feedback and strong technical support.
We also thank the anonymous reviewers for their constructive feedback.
%\kenneth{Who else also helped with data labeling?} \edward{cha cha}
%We also thank our collaborators at Adobe Research for their valuable feedback.
%\kenneth{We should probably include Xin and name names.}
%We would like to thanks the Huck Institutes of the Life Sciences' Coronavirus Research Seed Fund (CRSF) and the College of IST COVID-19 Seed Fund at Penn State University who support the construction of CODA-19.
%We also thank Tiffany Knearem for the feedback for designing word cloud visualization and workers who participated the human evaluation study.
This research was partially supported by the Seed
Grant~(2020) from the College of Information Sciences and
Technology~(IST), Pennsylvania State University.

%\paragraph{Dataset Collection Process and Conditions.}

\bibliography{anthology}
\bibliographystyle{acl_natbib}

% \appendix
% \section{Appendix}
% \label{sec:appendix}
% \input{appendix.tex}

\end{document}